\documentclass[journal]{IEEEtran}
\ifCLASSINFOpdf
\else
\fi
\usepackage{graphics}
\usepackage{subfigure}
\usepackage{epsfig}
\usepackage{multirow,rotating}
\usepackage{amsmath}
\usepackage{array}
\usepackage{multirow}
\usepackage{colortbl}
\usepackage{cite}
\usepackage{url}
\usepackage{times}
\usepackage{amssymb}
\usepackage{subfigure}
\usepackage{tabularx}
\usepackage{algorithm,algpseudocode}
\usepackage{mathtools}
\usepackage{amssymb}
\usepackage{pifont}

\usepackage{booktabs}
\usepackage{pdflscape}
\usepackage{epstopdf}
\usepackage{mathrsfs}
\usepackage{caption}
\usepackage{makecell}
\usepackage{tabularx,booktabs}
\usepackage{graphicx}
\usepackage{changes}
\algdef{SE}[DOWHILE]{Do}{doWhile}{\algorithmicdo}[1]{\algorithmicwhile\ #1}%

\newcolumntype{C}{>{\centering\arraybackslash}X} 
\begin{document}
%
\title{Person Re-identification in Videos by Analyzing Spatio-Temporal Tubes}
%
%
%
  \author{
       Sk. Arif Ahmed,~\IEEEmembership{Member,~IEEE,}~Debi Prosad Dogra,~\IEEEmembership{Member,~IEEE,}~Heeseung Choi, Seungho Chae and Ig-Jae Kim
\thanks{Sk. Arif Ahmed (Email: arif.1984.in@ieee.org) is with NIT Durgapur, India}
\thanks{Debi Prosad Dogra (Email: dpdogra@iitbbs.ac.in) is with  IIT Bhubaneswar, India}
\thanks{Heeseung Choi (Email:hschoi@kist.re.kr) Seungho Chae (Email: seungho.chae@kist.re.kr) and Ig-Jae Kim (Email: drjay@kist.re.kr) are with KIST South Korea}
\    }
\maketitle


\begin{abstract}
Typical person re-identification frameworks search for $k$ best matches in a gallery of images that are often collected in varying conditions. The gallery may contain image sequences when re-identification is done on videos.  However, such a process is time consuming as re-identification has to be carried out multiple times. In this paper, we extract spatio-temporal sequences of frames (referred to as tubes) of moving persons and  apply a multi-stage processing to match a given query tube with a  gallery of  stored tubes recorded through other cameras. Initially, we apply a binary classifier to remove noisy images from the input query tube. In the next step, we use a key-pose detection-based query minimization. This reduces the length of the query tube by removing redundant frames. Finally, a 3-stage hierarchical re-identification framework is used to rank the output tubes as per the matching scores. Experiments with publicly available video re-identification datasets reveal that our framework is better than state-of-the-art methods. It ranks the tubes with an increased CMC accuracy of 6-8\% across multiple datasets.  Also, our method significantly reduces the number of false positives. A new video re-identification dataset, named Tube-based Re-identification Video Dataset (TRiViD), has been prepared with an aim to help the re-identification research community.

\end{abstract}

\begin{IEEEkeywords}
Trajectory analysis, anomaly detection, ELM, HTM, bio-inspired leaarning
\end{IEEEkeywords}
\section{Introduction}
Person re-identification (Re-Id) is useful in various intelligent video surveillance applications. The task can be considered as image retrieval problem, where a query image of a person (probe) is given and we search the person in a set of images extracted from different cameras (gallery). The query can be a single image~\cite{sun2017svdnet} or multiple images~\cite{deng2018image}. Often multi-image query uses early fusion of images and generate an average query image~\cite{xu2017jointly}. The method thus consumes higher computational power as compared to single image-based methods. Advanced hardware and efficient learning frameworks have encouraged the researchers to focus on designing Re-Id systems applicable to videos. However, video-based re-identification research is still in its infancy~\cite{lv2018unsupervised,chen2018video}. Even though the existing video Re-Id applications seem to be promising, such methods often fail in low resolution videos, crowded environment, or in the presence of significant camera  angle variations. It has also been observed that the query image or video has to be selected judiciously to obtain good retrieval results. Choosing an improper image or video may lead to poor quality of retrieval. In this paper, we detect and track humans in movement and construct spatio-temporal tubes that are used in the re-identification framework. We also propose a method for selecting optimum set of key pose images and use a 3-stage learning framework to re-identify persons appearing in different cameras. To accomplish this, we have made the following contributions in this paper:
\begin{itemize}
\item We propose a learning-based method to select an optimum set of key pose images to reconstruct the query tube by minimizing its length in terms of number of frames.
\item We propose a 3-stage hierarchical framework that has been built using  (i) SVDNet guided Re-Id architecture, (ii) self-similarity estimation, and (iii) temporal correlation analysis to rank the tubes of the gallery.
\item We introduce a new video dataset, named Tube-based Re-identification Video Dataset (TRiViD) that has been prepared with an aim to help the re-identification research community.
\end{itemize}

Rest of the paper is organized as follows. In Section 2, we discuss the state-of-the-art of person re-identification research. Section 3 presents the proposed Re-Id framework with various components. Experiment results are presented in Section 4. Conclusion and future work are presented in Section 5.

\section{Related Work}
Person re-identification applications are growing rapidly in numbers. However, humongeous growth in CCTV surveillance has thrown up various challenges to the re-identification research community. The primary challenges are to handle large volume of data~\cite{zheng2016mars,zheng2015scalable}, tracking in complex environment~\cite{ristani2016performance,zheng2017person}, presence of group~\cite{chen2018group}, occlusion~\cite{huang2018adversarially}, varying pose and style across different cameras~\cite{deng2018image,liu2018pose,sarfraz2017pose,zhong2018camera}, etc. The process of Re-Id can be categorized as image-guided~\cite{barman2017shape,chang2018multi,chen2018group,deng2018image} and video-guided~\cite{lv2018unsupervised,chen2018video,chung2017two,wu2018exploit,zhang2017multi}. The image-guided methods typically use deep neural networks for feature representation and re-identification, whereas the video-guided methods typically use recurrent convolutional networks (RNN) to embed the temporal information such as optical flow~\cite{chung2017two}, sequence of pose, etc. Table~\ref{rel} summarizes recent progress in person re-identification. In recent years, late fusion of different scores~\cite{barman2017shape,paisitkriangkrai2015learning} has shown significant improvement over the final ranking. Our method is similar to a typical delayed or late fusion guided method. We refine search results obtained using convolutional neural networks with the help of temporal correlation analysis.

\begin{table*}[!htb]
\resizebox{18cm}{!}{%
\begin{tabular}{@{}lll@{}}
\toprule
\textbf{Reference} & \textbf{Method Overview}  \\ \midrule
Lv et al.~\cite{lv2018unsupervised}             & Motion and image based features Recurrent convolutional network for video-based person re-identification
                                \\
Barman et al.~\cite{barman2017shape}             & \begin{tabular}[c]{@{}l@{}}Graph theory and multiple algorithm fusion-based algorithm
SHaPE: A Novel Graph Theoretic Algorithm for Making Consensus-based Decisions in Person Re-identification Systems\end{tabular}

                               \\
 Chang et al.~\cite{chang2018multi}             & Visual appearance and multiple semantic level features Multi-Level Factorization Net for Person Re-Identification

                                \\
Chen et al.~\cite{chen2018group}             & \begin{tabular}[c]{@{}l@{}}Fusion of local similarity and group similarity-based DNN and CRF
Group Consistent Similarity Learning via Deep CRF for Person Re-Identification\end{tabular}

                                \\
Chen et al.~\cite{chen2018video}             & \begin{tabular}[c]{@{}l@{}}Divides a long person sequence into short snippet and match snippets for re-identification
Video Person Re-identification with Competitive Snippet-similarity Aggregation and Co-attentive Snippet Embedding\end{tabular}

                                \\
Chung et al.~\cite{chung2017two}             & \begin{tabular}[c]{@{}l@{}}Learn spatial and temporal similarity and used weighed fusion
A Two Stream Siamese Convolutional Neural Network For Person Re-Identification\end{tabular}

                                \\
Deng et al.~\cite{deng2018image}             & \begin{tabular}[c]{@{}l@{}}Learn self similarity and domain dissimilarity
Image-Image Domain Adaptation with Preserved Self-Similarity and Domain-Dissimilarity for Person Re-identification\end{tabular}


                                \\
 He et al.~\cite{he2018deep}             & \begin{tabular}[c]{@{}l@{}}Deep pixel-level CNN for person re-identification from partially observed images. Deep Spatial Feature Reconstruction for Partial Person Re-identification: Alignment-free Approach\end{tabular}

                               \\
 Huang et al.~\cite{huang2018adversarially}             & \begin{tabular}[c]{@{}l@{}}Proposed augmented training data generation for person re-identification. Adversarially Occluded Samples for Person Re-identification\end{tabular}

                                \\
Kalayeh et al.~\cite{kalayeh2018human}             & \begin{tabular}[c]{@{}l@{}}Proposed human semantic parts model to train state-of-the-art deep networks and calculate weighted average. Human Semantic Parsing for Person Re-identification\end{tabular}

                               \\
 Li et al.~\cite{li2018diversity}             & \begin{tabular}[c]{@{}l@{}}Distinct body parts-based attention model for re-identification. Diversity Regularized Spatiotemporal Attention for Video-based Person Re-identification\end{tabular}

                                \\
Li et al.~\cite{li2018harmonious}             & \begin{tabular}[c]{@{}l@{}}Harmonious attention network consists of pixel-level and bounding box level attention as feature. Harmonious Attention Network for Person Re-Identification\end{tabular}

                                \\
Liu et al.~\cite{liu2018pose}             & Augmented pose of persons and generate training set as used to re-identify persons Pose Transferrable Person Re-Identification

                                \\
Liu et al.~\cite{liu2017stepwise}             & \begin{tabular}[c]{@{}l@{}}Tracklets have been used as training and re-identification. Stepwise Metric Promotion for Unsupervised Video Person Re-identification\end{tabular}

                              \\
Lv et al.~\cite{lv2018unsupervised}             & \begin{tabular}[c]{@{}l@{}}Transfer learning have been used to learn spatio-temporal pattern in unsupervised manner. Unsupervised Cross-dataset Person Re-identification by Transfer Learning of Spatial-Temporal Patterns\end{tabular}

                               \\
Fu et al.~\cite{fu2017multi}             & Used multi-scale feature representation and chose correct scale for matching Multi-scale Deep Learning Architectures for Person Re-identification

                               \\
Tomasi et al.~\cite{tomasi2018features}             & Proposed method for selection of good features for re-identification Features for Multi-Target Multi-Camera Tracking and Re-Identification

                               \\
 Roy et al.~\cite{roy2018exploiting}             & \begin{tabular}[c]{@{}l@{}}Minimized the labeling effort by choosing minimum image for labeling task in re-identification. Exploiting Transitivity for Learning Person Re-identification Models on a Budget\end{tabular}

                               \\
Sarfraz et al.~\cite{sarfraz2017pose}             & \begin{tabular}[c]{@{}l@{}}Used fine and coarse pose information for deep re-identification. A Pose-Sensitive Embedding for Person Re-Identification with Expanded Cross Neighborhood Re-Ranking\end{tabular}
                               \\
Shen et al.~\cite{shen2018deep}             & \begin{tabular}[c]{@{}l@{}}Proposed group-shuffling random walk network for fully utilizing train and test images. Deep Group-shuffling Random Walk for Person Re-identification \end{tabular}

                                \\
 Shen et al.~\cite{shen2018end}             & \begin{tabular}[c]{@{}l@{}}Proposed Kronecker Product Matching module to match feature maps of different persons in an end-to-end trainable deep neural network. End-to-End Deep Kronecker-Product Matching for Person Re-identification \end{tabular}

                                 \\
 Si et al.~\cite{si2018dual}             & \begin{tabular}[c]{@{}l@{}}Uses and learn context-aware feature sequences and perform attentive sequence comparison simultaneously. Dual Attention Matching Network for Context-Aware Feature Sequence based Person Re-Identification
\end{tabular}
                                 \\
Wang et al.~\cite{wang2018person}             & \begin{tabular}[c]{@{}l@{}}Deep architecture named BraidNet is proposed. It uses the cascaded Wconv structure learns to extract the comparison features Images. Person Re-identification with Cascaded Pairwise Convolutions\end{tabular}
                                 \\
 Wu et al.~\cite{wu2018exploit}             & \begin{tabular}[c]{@{}l@{}}It propose an approach to exploiting unsupervised Convolutional Neural Network (CNN) feature representation via stepwise learning. Exploit the Unknown Gradually: One-Shot Video-Based Person Re-Identification by Stepwise Learning\end{tabular}

                               \\
 Xu et al.~\cite{xu2018attention}             & Body parts-based attention network for re-identification Attention-Aware Compositional Network for Person Re-identification

                                \\
Xu et al.~\cite{xu2017jointly}             &\begin{tabular}[c]{@{}l@{}}Joint Spatial and Temporal Attention Pooling Network (ASTPN) has been used in video sequences. Jointly Attentive Spatial-Temporal Pooling Networks for Video-based Person Re-Identification\end{tabular}

                                \\
Zhang et al.~\cite{zhang2017multi}             & Sequential decision making has been used to identify each frame in a video Multi-shot Pedestrian Re-identification via Sequential Decision Making

                                 \\
Zhong et al.~\cite{zhong2018camera}             & Used style transfer across different camera to improve re-identification Camera Style Adaptation for Person Re-identification

                                \\

      \bottomrule
\end{tabular}
}
\caption{Recent progress in person re-identification research}
\label{rel}
\end{table*}

\begin{figure*}[!htb]
\begin{center}
\includegraphics[scale=0.5]{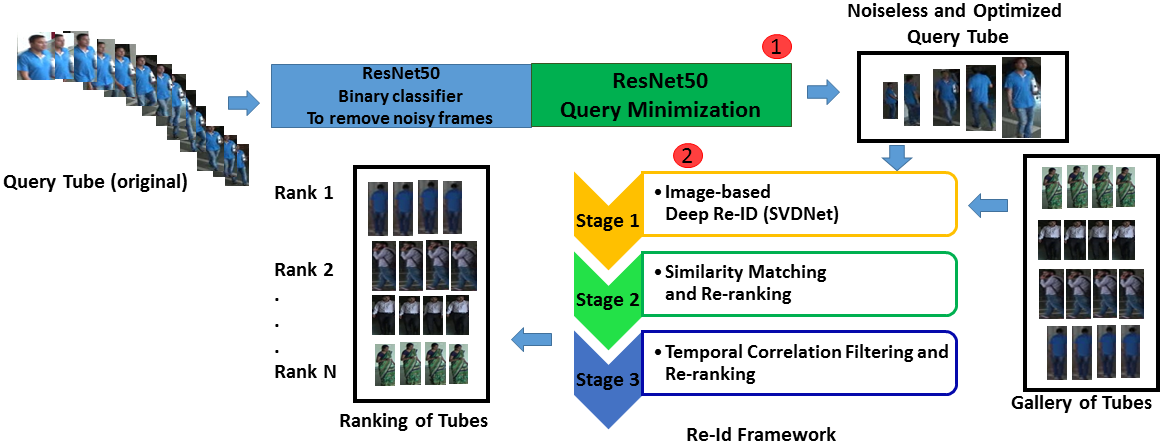}
\end{center}
\caption{The proposed method for Tube-to-tube Re-identification. Our contributions are marked with circle. The method takes a tube as query and rank the tubes by best matching.}
\label{fig:proposed}
\end{figure*}
\section{Proposed Approach}
Our method can be regarded as tracking followed by re-identification. Moving persons are tracked using Simple Online Deep Tracking (SODT) that has been developed using YOLO~\cite{jensen2017evaluating} framework. A tube is defined as the sequence of spatio-temporal frames of a moving person. Training is done using the videos captured by a camera. Videos captured using cameras are used to construct the gallery of tubes. Assume a gallery ($G$) contains $n$ tubes as given in~(\ref{eq:1}).
\begin{equation}
\label{eq:1}
G=\{T_1,T_2,T_3,...,T_n\}
\end{equation}
Suppose a tube ($T$) in the gallery contains $m$ frames as given in~(\ref{eq:2}).
\begin{equation}
\label{eq:2}
T=\{I_1,I_2,I_3,...,I_m\}
\end{equation}

At the time of re-identification, a query tube is given as a probe. First, the noisy frames are eliminated and the query tube is minimized. Next, frames of the revised query tube are passed through a 3-stage hierarchical re-ranking process to get the final ranking of the tubes in the gallery. The method is depicted in Figure~\ref{fig:proposed}.

\subsection{Query Minimization}
\begin{figure*}[!htb]
\begin{center}
\includegraphics[scale=0.5]{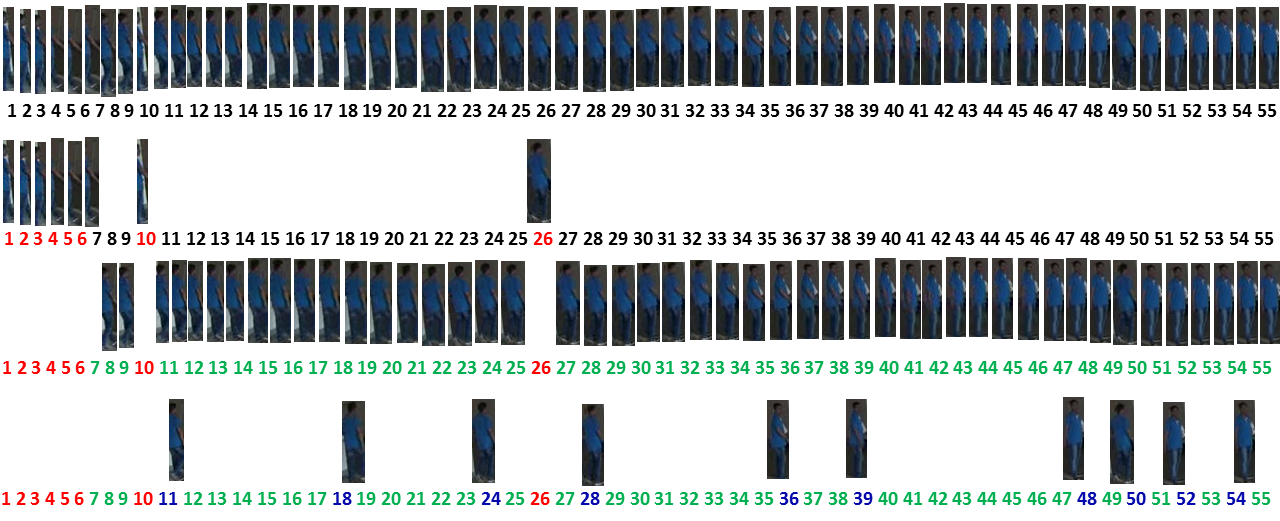}
\end{center}
\caption{Examples of original tube (first row), detected noisy frames (second row), tube after noise removal (third row), and minimized tube for query execution (fourth row) taken from the TRiViD dataset.}
\label{fig:noise}
\end{figure*}

Re-identification using multiple images usually performs better as compared to single image-based frameworks. However, the former method consumes more computational power. Also, selecting a set of frames that can uniquely represent a tube can be challenging.  To address this, we have used a deep similarity matching architecture to select a set of representative frames based on pose dissimilarity. First, a query tube is passed through binary classifier to remove noisy frames such as blurry, cropped, low-quality, etc. Next, a ResNet50~\cite{he2016deep} framework has been trained using a few query tubes containing similar looking images. The similarity cost $(\sigma_{ij})$ is calculated using~(\ref{eq:sim}).
\begin{equation}
\label{eq:sim}
    \sigma_{ij}=ResNet50(I_j,I_k)
\end{equation}
The input tube contains $m$ images, whereas the output query tube contains $n$ images such that $n<<m$. The images in the optimized query tube can be represented using~(\ref{eq:3}).
\begin{equation}
\label{eq:3}
    Q=\{I_1,I_2,I_3,...,I_n\}
\end{equation}
The pairwise query cost function $(\xi)$ for a given frame $(I_i)$ and other frame $(I_j)$ is defined in~(\ref{eq:cost1}).

\begin{equation}
\label{eq:cost1}
    \xi_{ij}=\min(\sigma_{ij}), \forall j
\end{equation}
The loss of energy is defined as given in~(\ref{eq:cost2}).
\begin{equation}
\label{eq:cost2}
    \gamma_{ij}=\forall j, \max (\sigma_{ij})
\end{equation}
The optimal query energy ($E$) is defined in~(\ref{eq:cost3}), where $\hat{Q}$ is the set of images that are not included in $Q$ and $\phi$ is a weighting parameter called query threshold (between 0-1). Larger $\phi$ produces higher number of images in $Q$.

\begin{equation}
\label{eq:cost3}
    E=\sum_{i,j \in Q}{\phi \xi_{ij}}+\sum_{i \in Q, j \in \hat{Q}}{\gamma_{ij}}
\end{equation}
Figure~\ref{fig:noise} depicts the steps and the minimized query images TRiViD dataset.

\subsection{Image Re-identification using SVDNet}
\begin{figure}[!htb]
\begin{center}
\includegraphics[scale=0.35]{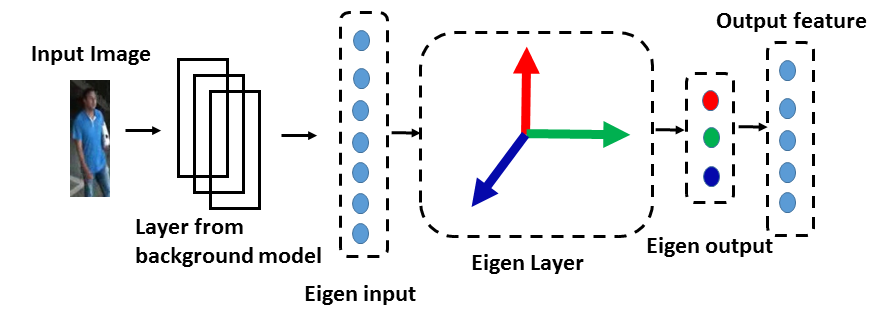}
\end{center}
\caption{Architecture of the SVDNet used in the fist stage of the re-identification framework shown in Figure 1. It contains an Eigenlayer before the fully connected layer. The Eigenlayer contains the weights to be used during training.}
\label{fig:svd}
\end{figure}

Our proposed method uses single image-based re-identification at the top layer of the hierarchy. We have used Singular Vector
Decomposition Network (SVDNet)~\cite{sun2017svdnet} as the baseline. It uses a convolutional neural network and an eigenlayer before the fully connected layer. The eigenlayer consists of a set of weights. Figure~\ref{fig:svd} demonstrates the architecture of a typical SVDNet. The outputs of SVDNet are a set of retrieved images with ranks up to $k$ as given in~(\ref{eq:svd}).

\begin{equation}
\label{eq:svd}
\textit{SVD}=\{I_1,I_2,I_3,...,I_k\}
\end{equation}

\subsection{Self Similarity Guided Re-ranking}
\begin{figure*}[!htb]
\begin{center}
\includegraphics[scale=0.6]{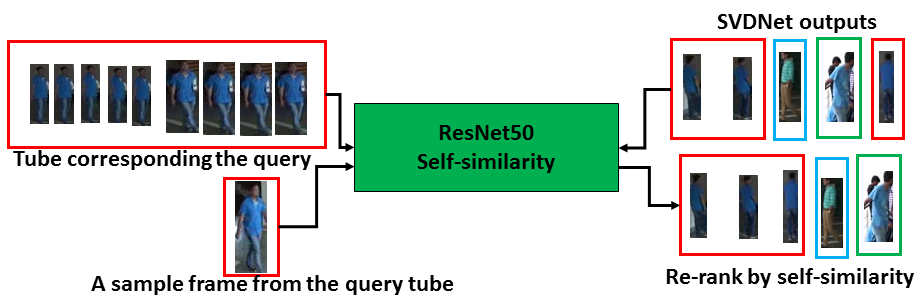}
\end{center}
\caption{The self similarity estimation layer. It learns to measure self-similarity during training. We use ResNet50~\cite{he2016deep} as the baseline. It takes a set of ranked images (SVDNet outputs) and produces a set of ranked images by introducing self-similarities between the query image and the retrieved images.}
\label{fig:self}
\end{figure*}

In the next step, we have aggregated the self-similarity scores with the SVDNet outputs. A typical ResNet50~\cite{he2016deep} architecture has been trained to learn self-similarity scores using the tubes of the query set. We assume the images available in a tube are similar. Next, a similarity score between the query image and every output image of SVD network up to rank $k$, is calculated. Finally, the scores are averaged and the images are re-ranked. This step ensures that the dissimilar images get pushed toward the end of the ranked sequence of the retrieved images. Figure~\ref{fig:self} illustrates this method.

\subsection{Tube Ranking by Temporal Correlation}
\begin{figure*}[!htb]
\begin{center}
\includegraphics[scale=0.45]{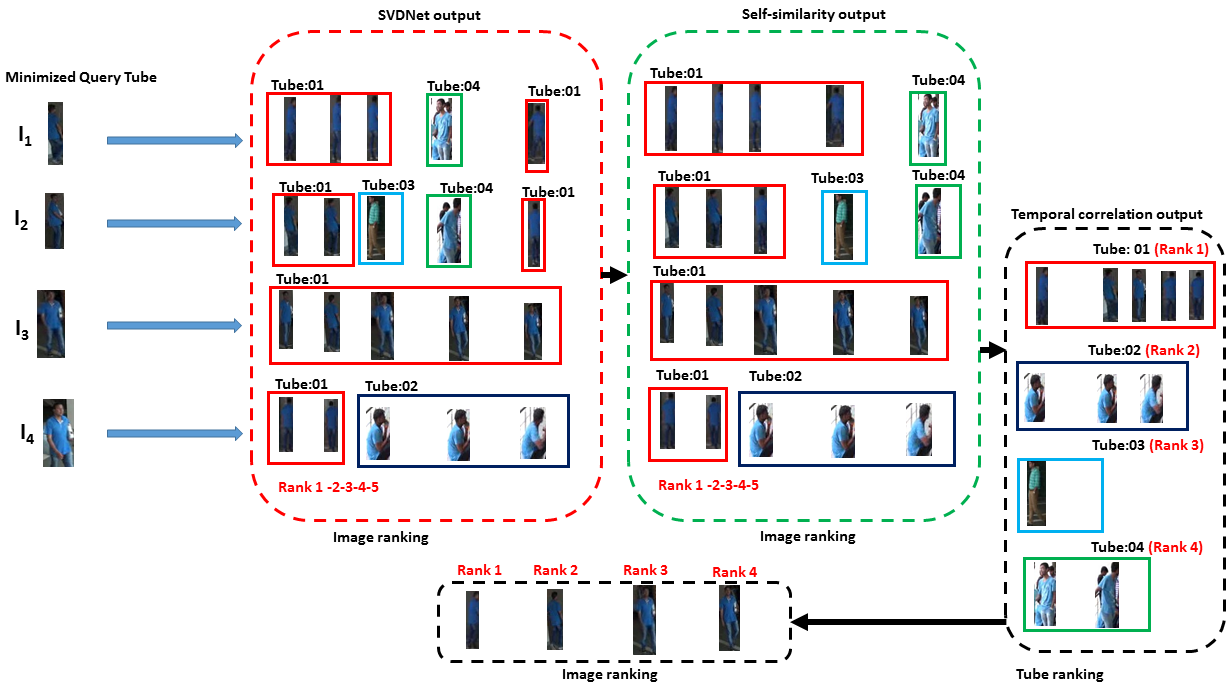}
\end{center}
\caption{Explanation of re-identification framework with the help of the proposed 3-stage framework depicted in Figure~\ref{fig:proposed}.}
\label{fig:temporal}
\end{figure*}

Final step of the proposed method is to rank the tubes by temporal correlation among the retrieved images. We assume the images that belong to a single tube, are temporally correlated as they are extracted by detection and tracking.
Let the result matrix up to rank $k$ for the query tube $Q$ after the first two stages be denoted by $R$. Weight of an image of $R$ can be estimated using~(\ref{eq:weight}).
\[
R=
\begin{bmatrix}
    I_{11}       & I_{12} & I_{13} & \dots & I_{1k} \\
    I_{21}       & I_{22} & I_{23} & \dots & I_{2k} \\
    \hdotsfor{5} \\
    I_{j1}       & I_{j2} & I_{j3} & \dots & I_{jk}
\end{bmatrix}
\]

\begin{equation}
\label{eq:weight}
\alpha_{I_{jk}}=\frac{1}{r}\text{, where $r$ is the rank of } I_{jk}
\end{equation}
Similarly, weight of a tube ($T_n$) can be estimated using~(\ref{eq:tweight}).
\begin{equation}
\label{eq:tweight}
\beta_{T_n} =\frac{\text{\# of images of $T_n$ that are in } R}{{\text{maximum \# of images belong to } T_n \in R}}
\end{equation}
Finally, the temporal correlation cost ($\tau_{I_{jk}}$) of an image in $R$ can be estimated as given in~(\ref{eq:cor}).
\begin{equation}
\label{eq:cor}
\tau_{I_{jk}}=\alpha_{I_{jk}} \times \beta_{T_n}, \text{for} \:I_{jk} \in T_n
\end{equation}
Based on the temporal correlation, the retrieved tubes are ranked. Let the ranked tubes up to $k$ be represented using~(\ref{eq:rtube}), where higher rank tubes have higher weights.
\begin{equation}
\label{eq:rtube}
R_{tube}=\{T_1,T_2,...,T_k\}
\end{equation}
The final ranked images are extracted by taking the highest scoring images from the tubes. The final ranked images are given in~(\ref{eq:tube2image}). Figure~\ref{fig:temporal} explains the whole process of tube ranking and selection of final set of frames.
\begin{equation}
\label{eq:tube2image}
R_{image}=\{I_1,I_2,...,I_l\}\text{, where }I_i \in T_i
\end{equation}

\section{Experiments}
We have evaluated our proposed approach on two public
datasets, iLIDS-VID~\cite{wang2014person} and PRID-11~\cite{hirzer2011person} that are often used for testing video-based re-identification frameworks. In addition to that, we have also prepared a new re-identification dataset. It has been recorded using 2 cameras in an indoor environment with human movements with moderately dense crowd (with more than 10 people appearing within 4-6 sq-mt), varying camera angles, and persons with similar clothing. Such situations have not been covered yet in existing re-identification video datasets. Details about these datasets are presented in Table~\ref{tbl:dataset}. Several experiments have been conducted to validate our method and a through comparative analysis has been performed.

\begin{table}[!htb]
\resizebox{\columnwidth}{!}{%
\begin{tabular}{@{}lllll@{}}
\toprule
\multicolumn{1}{c}{\textbf{Dataset}} & \multicolumn{1}{c}{\textbf{\begin{tabular}[c]{@{}l@{}}Number of\\ Camera\end{tabular}}} & \multicolumn{1}{c}{\textbf{\begin{tabular}[c]{@{}l@{}}Person\\ Re-appeared\end{tabular}}} & \multicolumn{1}{c}{\textbf{\begin{tabular}[c]{@{}l@{}}Gallery \\ Size\end{tabular}}} & \textbf{Challenges} \\ \midrule
PRID-11~\cite{hirzer2011person}                        & 2                                                                                       & 245                                                                                        & 475                                                                                  & Large volume        \\
iLIDS-VID~\cite{wang2014person}                     & 2                                                                                       & 119                                                                                        & 300                                                                                  & Clothing Similarity \\
TRiViD                             & 2                                                                                       & 47                                                                                         & 342                                                                                  & Dense, Tracking, Similarity            \\ \bottomrule
\end{tabular}
}
\caption{Dataset used in our experiments. Only TRiViD dataset is tracked to extract tube. In other datset the given sequence of images are considered as tube}
\label{tbl:dataset}
\end{table}
\textbf{Evaluation Metrics and Strategy:} We have followed the well known experimental protocols for evaluating the method. For iLIDS-VID and TRiViD dataset videos, the tubes are randomly split into 50\% for training and 50\% for testing. For PRID-11, we have followed the experimental setup as proposed in~\cite{mclaughlin2016recurrent,wang2014person,xu2017jointly,zhou2017see,chen2018video}.
Only first 200 persons who appeared in both cameras of the PRID-11 dataset, have been used in our experiments. A 10-folds cross validation scheme has been adopted and the average results are reported. We have prepared Cumulative Matching Characteristics (CMC) and mean average precision (mAP) curves to evaluate and compare the performance.

\subsection{Comparative Analysis}
\begin{figure}[!htb]
\begin{center}
\includegraphics[scale=0.5]{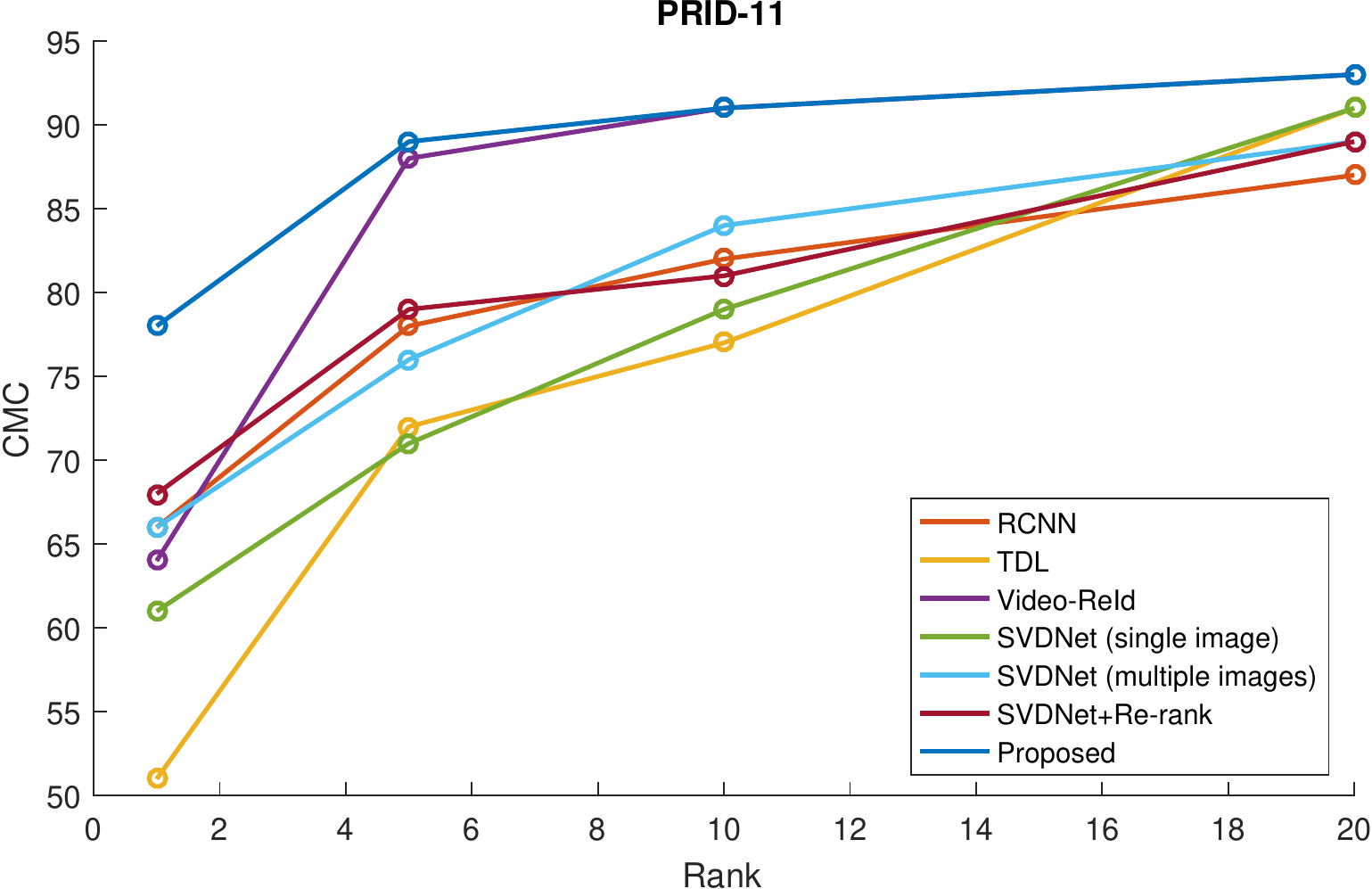}
\end{center}
\caption{The accuracy (CMC) in PRID-11 dataset using RCNN~\cite{mclaughlin2016recurrent}, TDL~\cite{you2016top}, Video re-id~\cite{mclaughlin2016recurrent}, SVDNet~\cite{sun2017svdnet} (single image), SVDNet (multiple images), SVDNet+Re-rank~\cite{paisitkriangkrai2015learning}.}
\label{fig:res1}
\end{figure}

\begin{figure}[!htb]
\begin{center}
\includegraphics[scale=0.5]{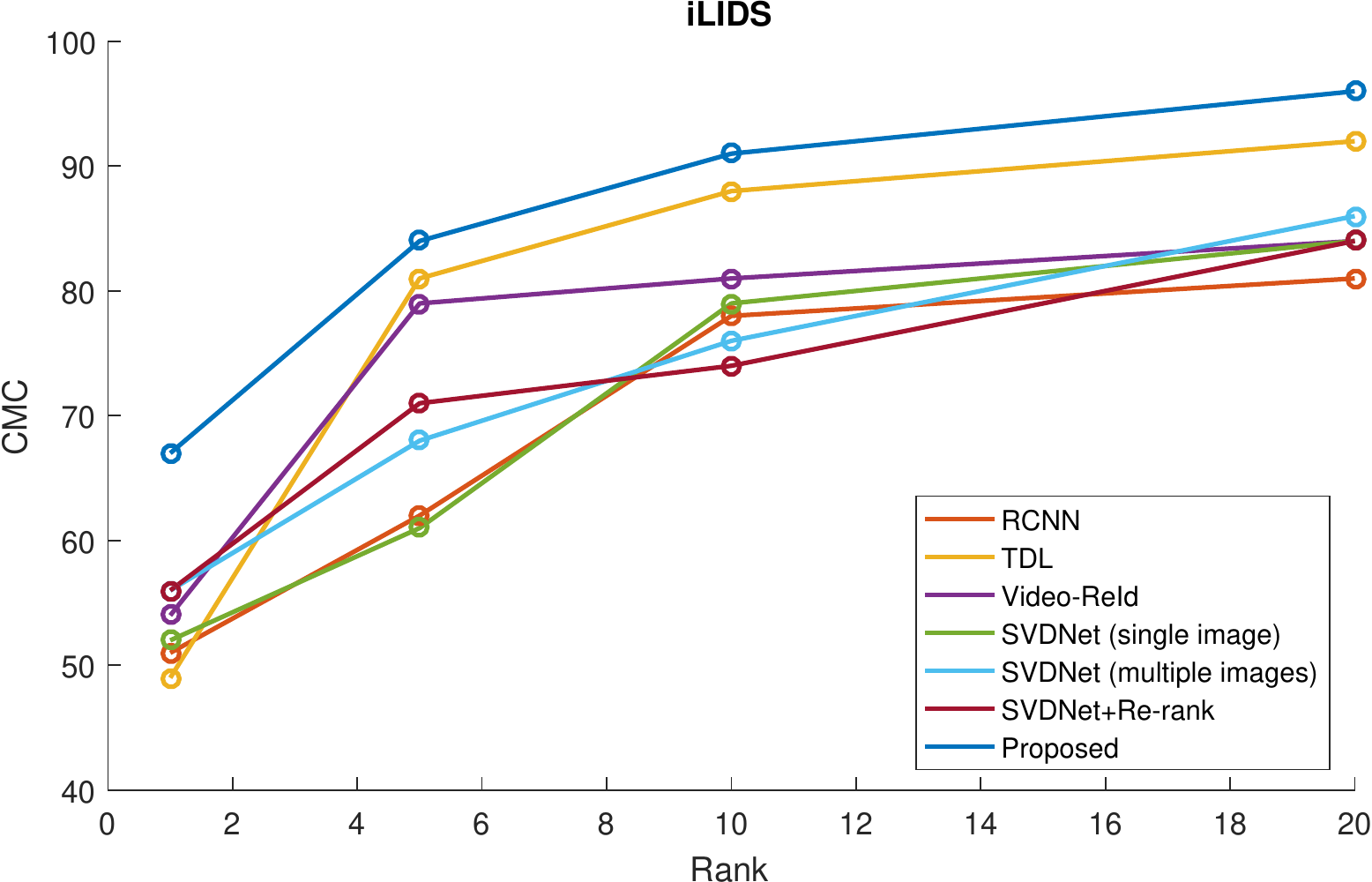}
\end{center}
\caption{The accuracy (CMC) in iLIDS dataset using RCNN~\cite{mclaughlin2016recurrent}, TDL~\cite{you2016top}, Video re-id~\cite{mclaughlin2016recurrent}, SVDNet~\cite{sun2017svdnet} (single image), SVDNet (multiple images), SVDNet+Re-rank~\cite{paisitkriangkrai2015learning}.}
\label{fig:res2}
\end{figure}

\begin{figure}[!htb]
\begin{center}
\includegraphics[scale=0.5]{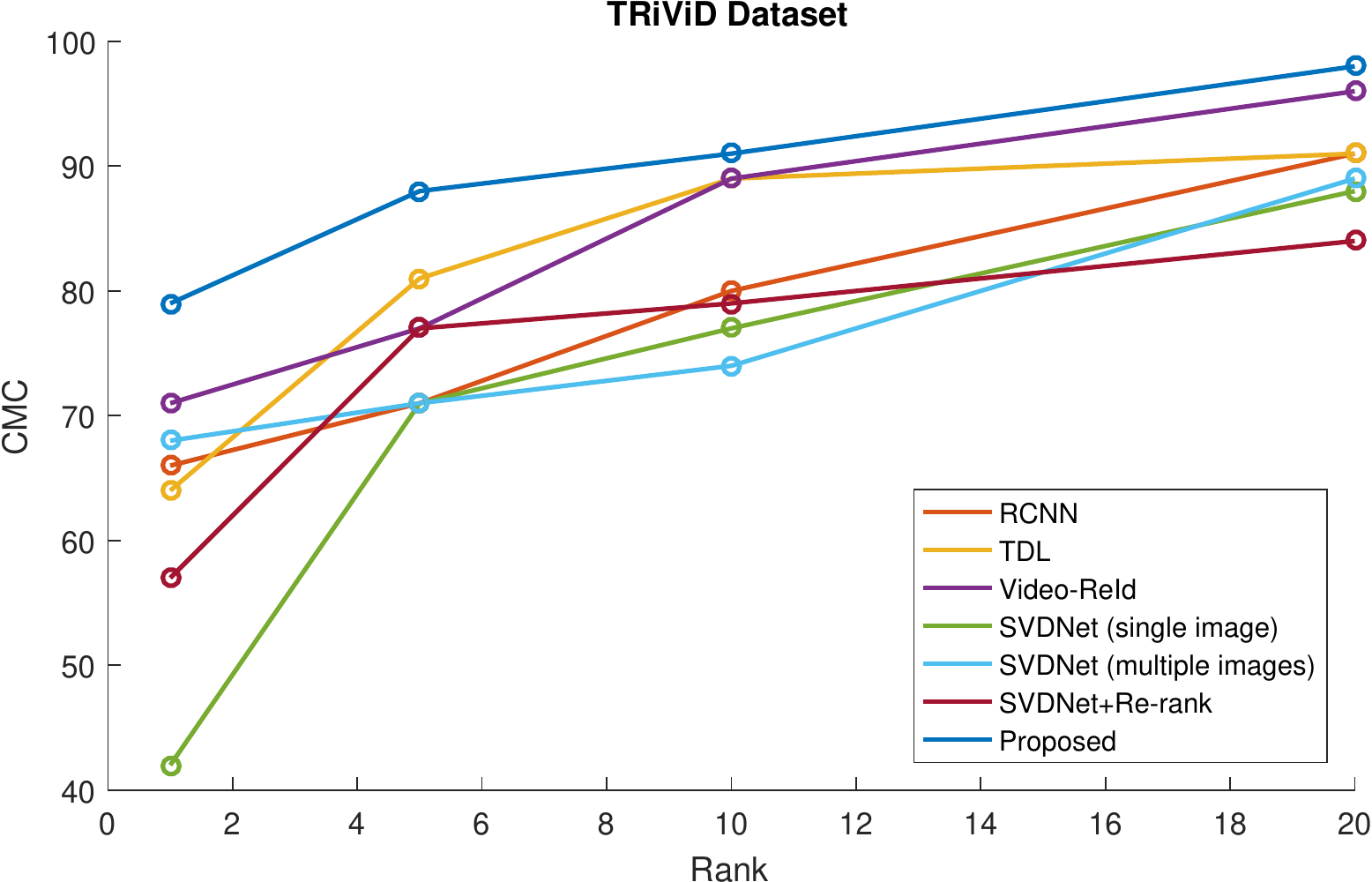}
\end{center}
\caption{The accuracy (CMC) using the TRiViD dataset with the help of RCNN~\cite{mclaughlin2016recurrent}, TDL~\cite{you2016top}, Video re-id~\cite{mclaughlin2016recurrent}, SVDNet~\cite{sun2017svdnet} (single image), SVDNet (multiple images), SVDNet+Re-rank~\cite{paisitkriangkrai2015learning}.}
\label{fig:res3}
\end{figure}

\begin{figure}[!htb]
\begin{center}
\includegraphics[scale=0.4]{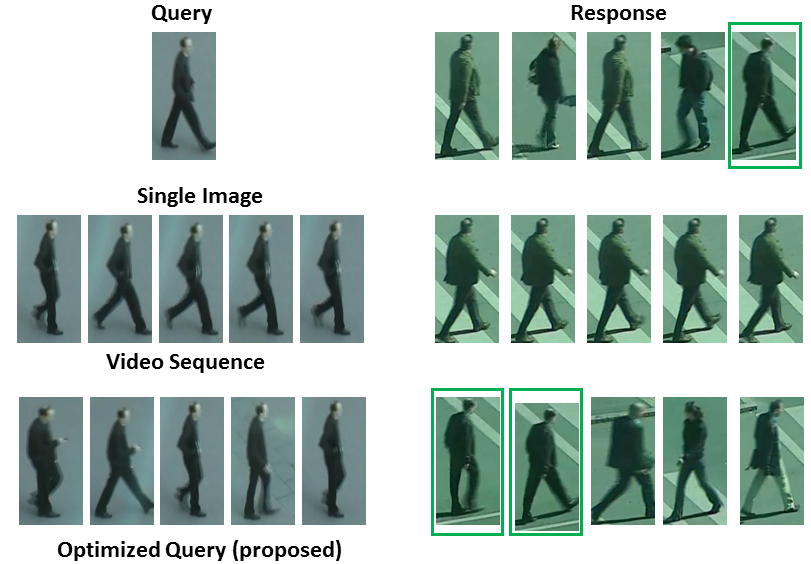}
\end{center}
\caption{Typical results obtained using PRID-11 dataset using single image query~\cite{sun2017svdnet}, video sequence~\cite{mclaughlin2016recurrent}, and using the proposed method. Green box indicates a correct retrieval.}
\label{fig:reid}
\end{figure}

As per the state-of-the-art, our work though unique in design has some similarities with video re-id methods proposed in~\cite{you2016top,mclaughlin2016recurrent}, multiple query-based method~\cite{sun2017svdnet}, and the re-ranking method~\cite{paisitkriangkrai2015learning}. Therefore, we have compared our approach with the above three recently proposed methods.
It has been observed that the proposed method can achieve a gain up to 9.6\% as compared to the state-of-the-art methods when top rank accuracy is estimated. Even if we compute the accuracy up to rank 20, our method has the upper hand with a margin of 3\%.
This is the USP of the proposed method and we claim it to be significant at this stage. This happens because our method tries to reduce the number of false positives which has not yet been addressed by the re-identification research community.
Figures~\ref{fig:res1}-\ref{fig:res3} represent CMC curves and Table~\ref{tbl:map} summarizes the mAP up to rank 20 across the three datasets. Figure~\ref{fig:reid} shows a typical query and response applied on PRID-11 dataset.


\begin{table}[!htb]
\centering
\label{tbl:map}
\begin{tabular}{@{}llll@{}}
\toprule
\textbf{Method/Dataset}   & \textbf{PRID} & \textbf{iLIDS} & \textbf{New} \\ \midrule
RCNN~\cite{mclaughlin2016recurrent}                      & 81.2             & 74.6           & 79.11        \\
TDL~\cite{you2016top}                      & 78.2             & 74.1           & 80           \\
Video-ReId~\cite{mclaughlin2016recurrent}                & 73.31            & 64.29          & 83.22        \\
SVD Net~\cite{sun2017svdnet} (Single Image)    & 76.44            & 69             & 79.11        \\
SVD Net~\cite{sun2017svdnet} (Multiple Images) & 79.21            & 66.71          & 82.66        \\
SVD Net+Re Rank~\cite{paisitkriangkrai2015learning}           & 77.25            & 69.2           & 78.6         \\
Proposed                  & 86.17            & 79.22          & 91.66        \\ \bottomrule
\end{tabular}
\caption{mAP (\%) up to rank 20 in across three video datasets}
\end{table}

\subsection{Computational Complexity Analysis}
\begin{figure*}[!htb]
\begin{center}
\includegraphics[scale=0.5]{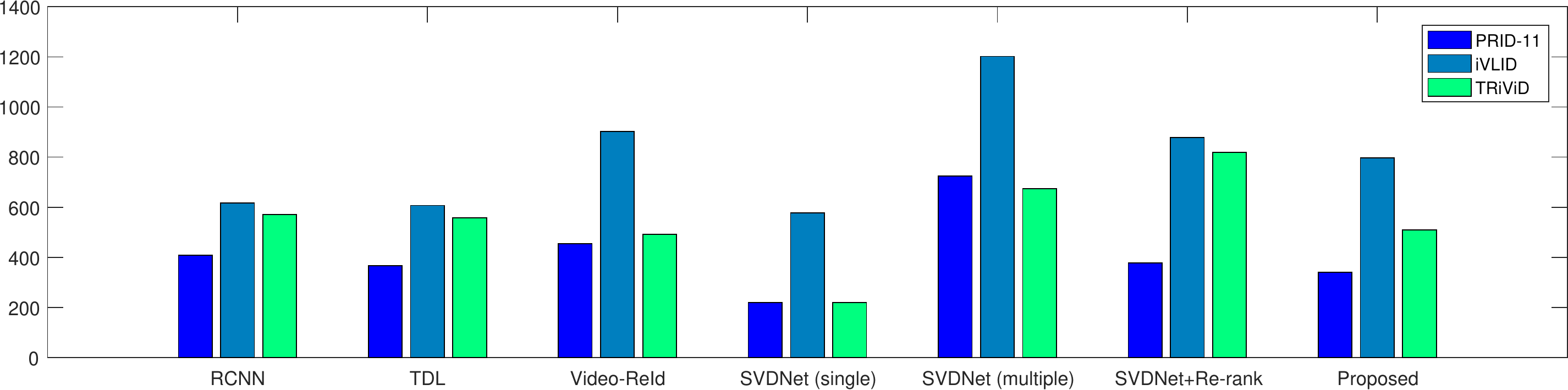}
\end{center}
\caption{Average response time (in seconds) for a given query by varying the datasets. We have taken 100 query tubes in random and calculated the average response time with the help of RCNN~\cite{mclaughlin2016recurrent}, TDL~\cite{you2016top}, Video re-id~\cite{mclaughlin2016recurrent}, SVDNet~\cite{sun2017svdnet} (single image), SVDNet (multiple images), SVDNet+Re-rank~\cite{paisitkriangkrai2015learning}.}
\label{fig:time}
\end{figure*}

re-identification in real-time is a challenging task. All research work carried out so far presume the gallery as a pre-recorded set of images and they try to rank best 5, 10, 15, 20 images from the set. However, executing a single query takes considerable time when multiple images are involvd in the query. We have carried out a comparative analysis on computation complexities across various re-identification frameworks including the proposed scheme. A Nvdia Quadro P5000 series GPU has been used to implement the frameworks. The results are reported in Figure~\ref{fig:time}. We have observed that the proposed tube-based re-identification framework takes lesser time as compared to video re-id framework proposed in~\cite{mclaughlin2016recurrent} and the multiple images-based re-id using SVDNet~\cite{sun2017svdnet}.

\subsection{Effect of $\phi$}
\begin{figure}[!htb]
\begin{center}
\includegraphics[scale=0.5]{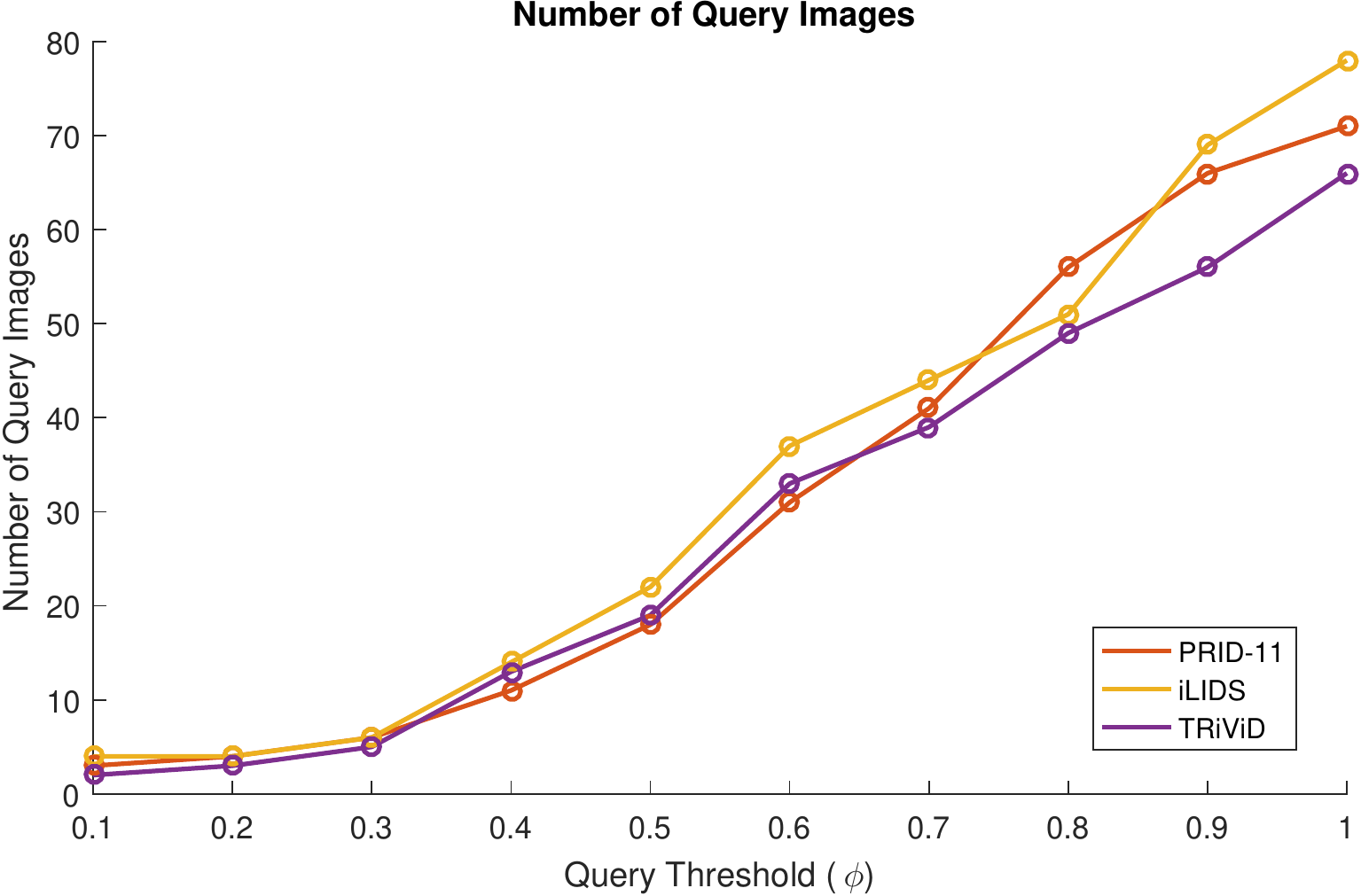}
\end{center}
\caption{Average number of query images by varying the query threshold $(\phi)$. We have taken 100 query sequences randomly and average number of optimized images, is reported. It may be observed that a higher $\phi$ produces more number of query images.}
\label{fig:q2}
\end{figure}

Our proposed method depends on the query threshold $(\phi)$. In this section, we present an analysis about the effect of $\phi$ on results. Figure~\ref{fig:q2} depicts the average number of query images generated from various query tubes. It may be observed that, higher $\phi$ produces more query images.

\begin{figure}[!htb]
\begin{center}
\includegraphics[scale=0.5]{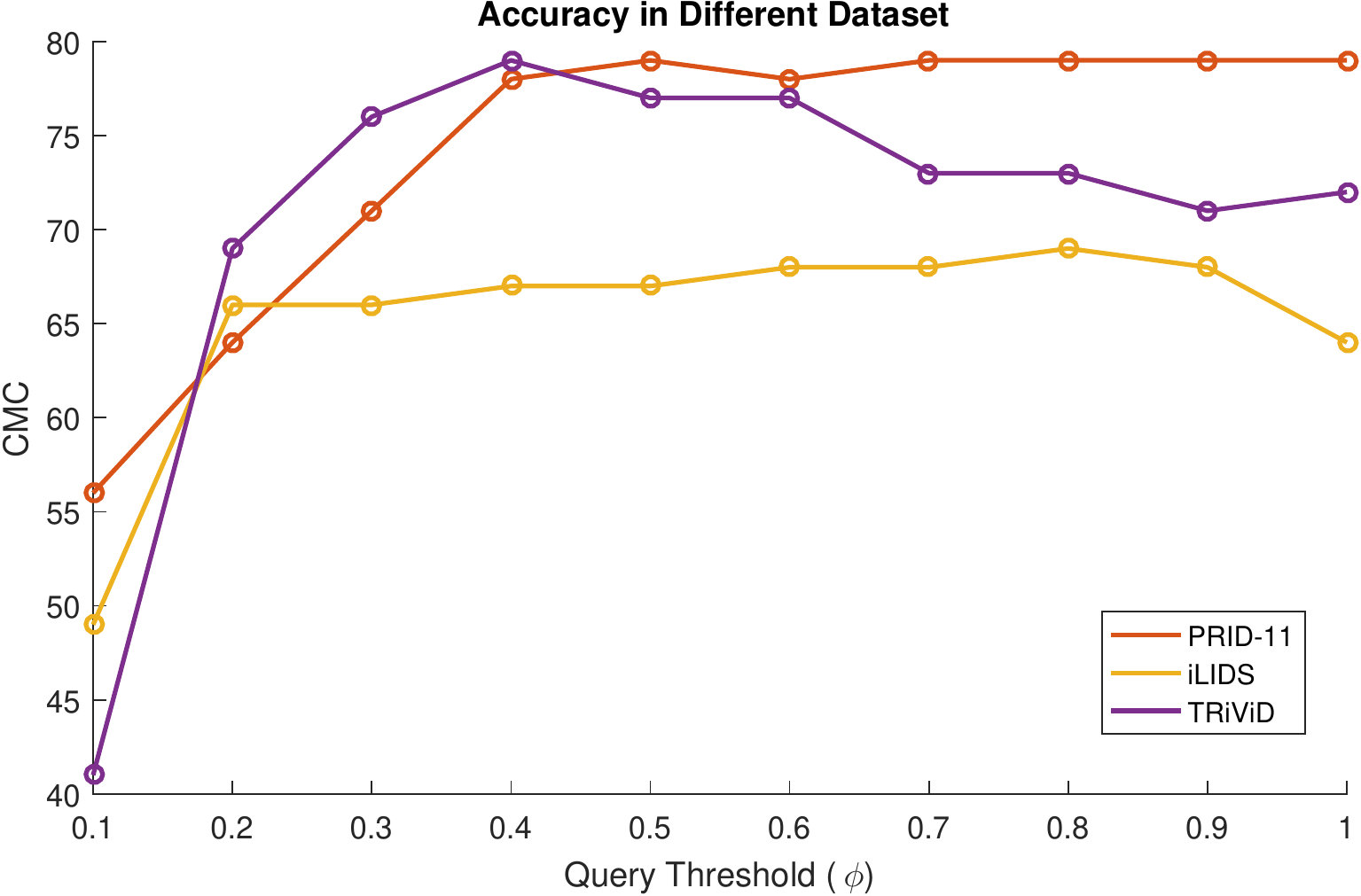}
\end{center}
\caption{Accuracy (CMC) by varying the query threshold $(\phi)$. We have taken 100 query sequences randomly and average is reported. It may be observed that a higher $\phi$ may not produce higher accuracy}
\label{fig:q3}
\end{figure}

Figure~\ref{fig:q3} depicts average CMC by varying $\phi$. It may be observed that the accuracy does not increase significantly when $\phi$ is increased above 0.4.

\begin{figure}[!htb]
\begin{center}
\includegraphics[scale=0.5]{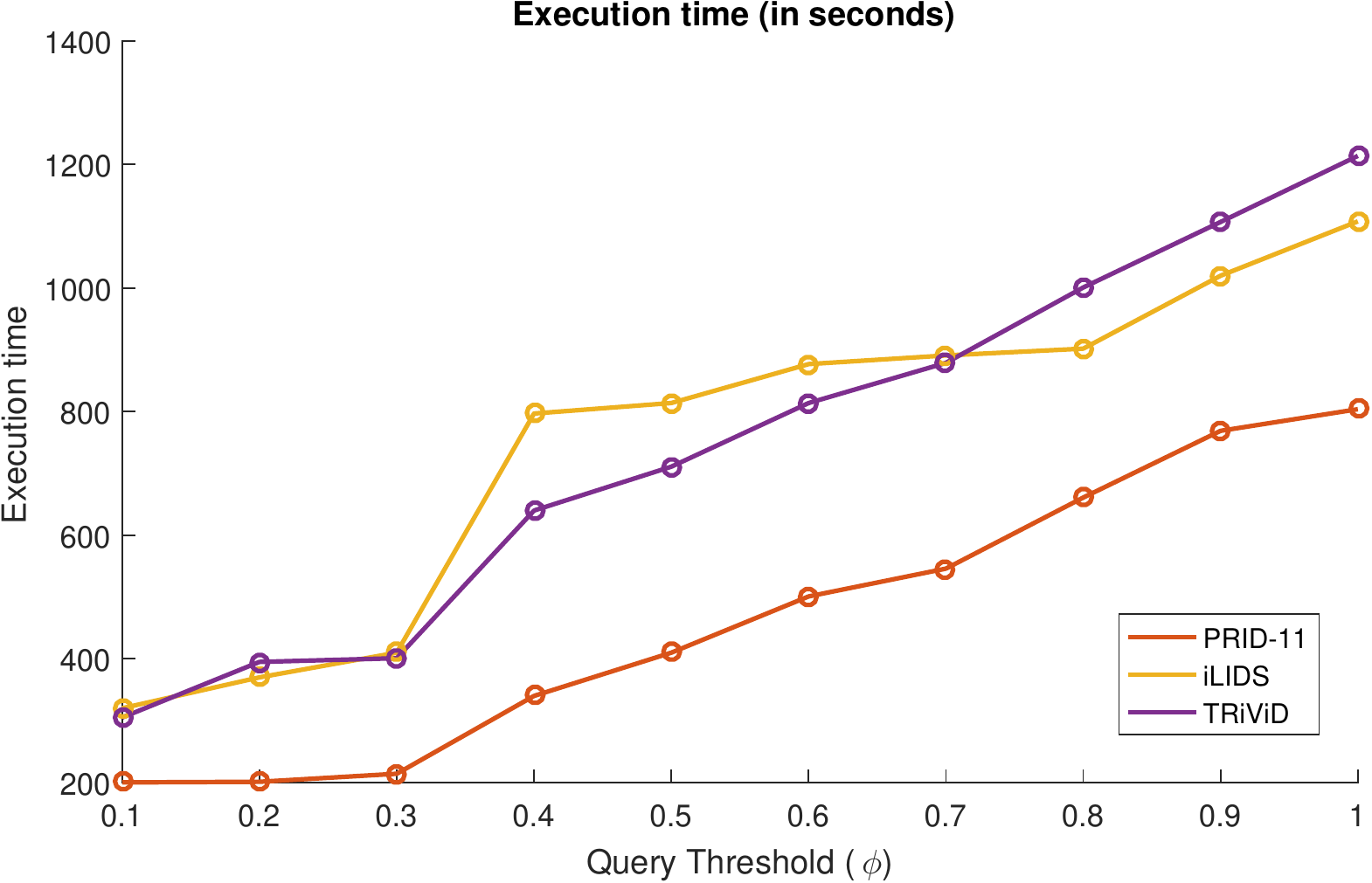}
\end{center}
\caption{Execution time by varying $\phi$. It may be observed that a higher $\phi$ takes more time to execute as it produces more query images.}
\label{fig:q4}
\end{figure}

Figure~\ref{fig:q4} presents execution time (in seconds) by varying the query threshold. It can also be observed that an increase in $\phi$ leads to higher response time. Therefore, we have used $\phi=0.4$ in our experiments.

\subsection{Results After Various Stages}
\begin{figure}[!htb]
\begin{center}
\includegraphics[scale=0.45]{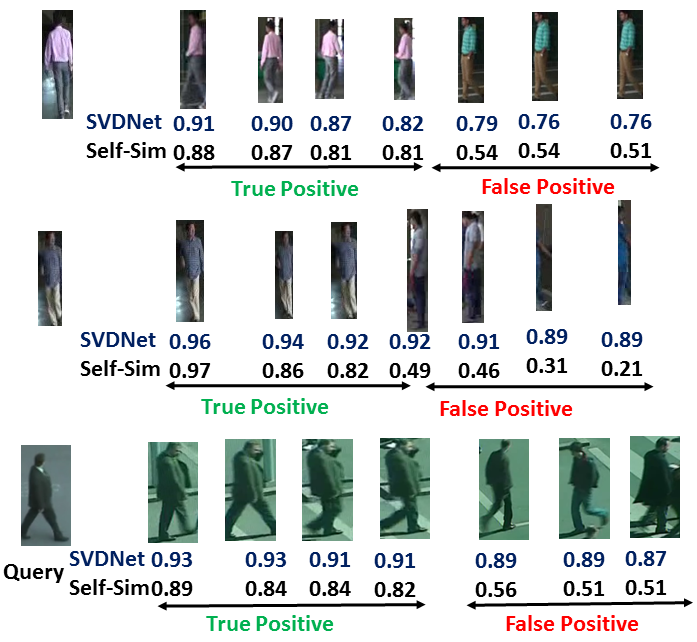}
\end{center}
\caption{Typical examples of SVDNet outputs and self-similarity scores in TRiViD (first two rows) and PRID-11~\cite{hirzer2011person} (last row).}
\label{fig:sim}
\end{figure}

In this section, we present the effect of various stages of the overall framework on re-identification results. Table~\ref{tbl:fusion} shows the accuracy (CMC) in each step of the proposed method. It may be observed that the proposed method gains 11\% rank-1 accuracy after the first stage and 7\% rank-1 accuracy after the second step. The method gains 7\% rank-20 accuracy in the first stage and 6\% rank-20 accuracy after the second stage. Table~\ref{tbl:fusion} shows the accuracy (CMC) in each step. Figure~\ref{fig:sim} shows an example of scores (true positives and false positives) during the self-similarity fusion. It may be observed that SVDNet output scores and similarity scores are high in case of true positives. Similarity scores are relatively low in case of false positives. More results can be found in the form of supplementary data.
\begin{table*}[!htb]
\center
\label{tbl:fusion}
\resizebox{15cm}{!}{%
\begin{tabular}{|l|l|l|l|l|l|l|l|l|l|l|l|l|}
\hline
\textbf{}                                                                                           & \multicolumn{4}{l|}{\textbf{PRID11~\cite{hirzer2011person}}}                & \multicolumn{4}{l|}{\textbf{iLIDS~\cite{wang2014person}}}                 & \multicolumn{4}{l|}{\textbf{TRiViD}}              \\ \hline
\textbf{Method/Top Rank}                                                                            & \textbf{1} & \textbf{5} & \textbf{10} & \textbf{20} & \textbf{1} & \textbf{5} & \textbf{10} & \textbf{20} & \textbf{1} & \textbf{5} & \textbf{10} & \textbf{20} \\ \hline
SVD Net (Multi Image)                                                                               & 66         & 76         & 84          & 89          & 56         & 68         & 76          & 86          & 68         & 71         & 74          & 89          \\ \hline
SVD Net+Self-similarity                                                                             & 69         & 77         & 84          & 89          & 61         & 71         & 79          & 86          & 71         & 77         & 76          & 91          \\ \hline
\begin{tabular}[c]{@{}l@{}}SVD Net+Self-similarity+ \\ Temporal Correlation (Proposed)\end{tabular} & 78         & 89         & 92          & 91          & 67         & 84         & 91          & 96          & 79         & 88         & 91          & 98          \\ \hline
\end{tabular}
}
\caption{Accuracy (CMC) in each step of the proposed method}
\end{table*}

\section{Conclusion}
In this paper, we propose a new person re-identification framework that is able to outperform existing re-identification schemes when applied on videos or sequence of frames.  The method uses a CNN-based framework (SVDNet) at the beginning. A self-similarity layer is used to refine the SVDNet scores. Finally, a temporal correlation layer is used to aggregate multiple query outputs and to match tubes. A query optimization has also been proposed to select an optimum set of images for a query tube. Our study reveals that the proposed method outperforms in several cases as compared to the state-of-the-art single image-based, multiple images-based, and video-based re-identification methods. The computational is also reasonably low.

One straight extension of the present work is to fuse methods like camera pose-based~\cite{deng2018image}, video-based~\cite{mclaughlin2016recurrent}, and description-based~\cite{chang2018multi}. It may lead to higher accuracy in complex situations. Also, group re-identification can be tried with the similar concept of tube guided analysis.

\section*{Acknowledgment}
The work has been funded under KIST Flagship Project (Project No.XXXX) and Global Knowledge Platform (GKP) of Indo-Korea Science and Technology Center (IKST) executed at IIT Bhubaneswar under the Project Code: XXX.
We gratefully acknowledge the support of NVIDIA Corporation with the donation of the Quadro P5000 GPU used for this research.

\bibliographystyle{IEEEtran}

\begin{IEEEbiography}[{\includegraphics[width=1in,height=1.25in,clip,keepaspectratio]{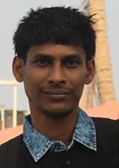}}]{Sk. Arif Ahmed}
has obtained a master degree in Computer Applications form West Bengal University of Technology. Currently working as an Assistant Professor at Haldia institute of Technology and Ph.D. candidate in NIT Durgapur, India. His area of interest is in the domain of computer vision, image processing, and scene analysis. He has already published 10 number of research articles in international journals and conferences.
\end{IEEEbiography}
\begin{IEEEbiography}[{\includegraphics[width=1in,height=1.25in,clip,keepaspectratio]{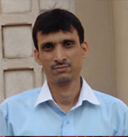}}]{Debi Prosad Dogra}
is an Assistant Professor in
the School of Electrical Sciences, IIT Bhubaneswar,
India. He received his M.Tech degree from IIT
Kanpur in 2003 after completing his B.Tech. (2001)
from HIT Haldia, India. After finishing his mas-
ters, he joined Haldia Institute of Technology as
a faculty members in the Department of Computer
Sc. \& Engineering (2003-2006). He has worked
with ETRI, South Korea during 2006-2007 as a
researcher. Dr. Dogra has published more than 75
international journal and conference papers in the
areas of computer vision, image segmentation, and healthcare analysis. He
is a member of IEEE.
\end{IEEEbiography}
\begin{IEEEbiography}[{\includegraphics[width=1in,height=1.25in,clip,keepaspectratio]{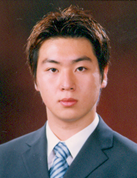}}]{Heeseung Choi}
Heeseung Choi received the B.S., M.S. and Ph.D. degrees in Electrical and Electronic Engineering from Yonsei University, Seoul, Korea, in 2004, 2006 and 2011 respectively. He had been a research member of BERC (Biometrics Engineering Research Center, Korea) and Computer Science and Engineering from Michigan State University, USA. He is currently a research member at Center for Imaging Media Research (CIMR) in Korea Institute of Science and Technology (KIST). His research interests include computer vision, biometrics, image processing, forensic science and pattern recognition.
\end{IEEEbiography}

\begin{IEEEbiography}[{\includegraphics[width=1in,height=1.25in,clip,keepaspectratio]{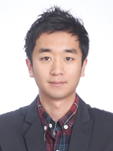}}]{Seungho Chae}
 received the PhD degree in computer science from Yonsei University, Seoul, Korea in 2018. Currently he is a Post-doc researcher in the Korea Institute of Science and Technology. His research interests lie in the fields of computer vision, augmented reality and human-computer interaction. Particularly, his research focuses on object tracking and person re-identification.
\end{IEEEbiography}
\begin{IEEEbiography}[{\includegraphics[width=1in,height=1.25in,clip,keepaspectratio]{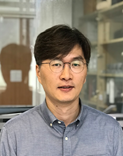}}]{Ig-Jae Kim}
is currently a Director of Center for Imaging Media Research, Korea Institute of Science and Technology (KIST), Seoul, South Korea. He is also associate professor at Korea University of Science and Technology. He received his Ph.D degree from EECS of Seoul National University in 2009, MS and BS degree from EE of Yonsei University, Seoul, South Korea, in 1998 and 1996 respectively. He had worked in Massachusetts Institute of Technology (MIT) Media Lab as a postdoctoral researcher (2009~2010). He has published over 80 fully-referred papers in international journal and conferences, including ACM Transaction on Graphics, SIGGRAPH, Pattern Recognition, ESWA, etc. He is interested in pattern recognition, computer vision, computer graphics, and computational photography.
\end{IEEEbiography}
\end{document}